\documentclass{article}

\usepackage{microtype}
\usepackage{graphicx}
\usepackage{caption} 
\usepackage{subcaption}
\usepackage{booktabs} 
\usepackage{tabularx}
\usepackage{pifont}
\newcommand{\na}{\textcolor{red}{\textbf{?}}}

\usepackage{hyperref}

\usepackage[accepted]{icml2025}

\usepackage{amsmath}
\usepackage{amssymb}
\usepackage{mathtools}
\usepackage{amsthm}
\usepackage{multirow}

\usepackage[capitalize,noabbrev]{cleveref}

\theoremstyle{plain}

\theoremstyle{definition}

\theoremstyle{remark}

\usepackage[textsize=tiny]{todonotes}
\usepackage{tcolorbox}

\newcommand{\dataname}[0]{\textsc{CounterMATH}}

\icmltitlerunning{\dataname: Counterexample-Driven Conceptual Reasoning in Mathematical LLMs}

\begin{document}

\twocolumn[
\icmltitle{\textit{One Example Shown, Many Concepts Known!} \\ Counterexample-Driven Conceptual Reasoning in Mathematical LLMs}



\icmlsetsymbol{equal}{*}
\icmlsetsymbol{corr}{$\dagger$}

\begin{icmlauthorlist}
\icmlauthor{Yinghui Li}{equal,tsinghua}
\icmlauthor{Jiayi Kuang}{equal,sysu}
\icmlauthor{Haojing Huang}{equal,tsinghua}
\icmlauthor{Zhikun Xu}{equal,corr,fudan,asu}
\icmlauthor{Xinnian Liang}{seed}
\icmlauthor{Yi Yu}{fudan}
\icmlauthor{Wenlian Lu}{fudan}
\icmlauthor{Yangning Li}{tsinghua,pcl}
\icmlauthor{Xiaoyu Tan}{corr,inf}
\icmlauthor{Chao Qu}{inf}
\icmlauthor{Ying Shen}{corr,sysu,FSIET}
\icmlauthor{Hai-Tao Zheng}{tsinghua,pcl}
\icmlauthor{Philip S. Yu}{uic}
\end{icmlauthorlist}

\icmlaffiliation{tsinghua}{Tsinghua University. E-mail: liyinghu20@mails.tsinghua.edu.cn}
\icmlaffiliation{sysu}{Sun-Yat Sen University}
\icmlaffiliation{fudan}{School of Mathematical Science, Fudan University}
\icmlaffiliation{seed}{Bytedance Inc.}
\icmlaffiliation{inf}{INFLY TECH (Shanghai) Co., Ltd.}
\icmlaffiliation{asu}{ARC Lab, Arizona State University}
\icmlaffiliation{uic}{University of Illinois Chicago}
\icmlaffiliation{pcl}{Peng Cheng Laboratory}
\icmlaffiliation{FSIET}{Guangdong Provincial Key Laboratory of Fire Science and Intelligent Emergency Technology, }

\icmlcorrespondingauthor{Ying Shen, Xiaoyu Tan, Zhikun Xu}{sheny76@mail.sysu.edu.cn, txywilliam1993@outlook.com, zhikunxu@asu.edu}

\icmlkeywords{Machine Learning, ICML}

\vskip 0.3in
]



\printAffiliationsAndNotice{\icmlEqualContribution} 

\begin{abstract}
Leveraging mathematical Large Language Models (LLMs) for proof generation is a fundamental topic in LLMs research. We argue that the ability of current LLMs to prove statements largely depends on whether they have encountered the relevant proof process during training. This reliance limits their deeper understanding of mathematical theorems and related concepts. Inspired by the pedagogical method of ``proof by counterexamples'' commonly used in human mathematics education, our work aims to enhance LLMs’ ability to conduct mathematical reasoning and proof through counterexamples. Specifically, we manually create a high-quality, university-level mathematical benchmark, \dataname{}, which requires LLMs to prove mathematical statements by providing counterexamples, thereby assessing their grasp of mathematical concepts. Additionally, we develop a data engineering framework to automatically obtain training data for further model improvement. Extensive experiments and detailed analyses demonstrate that \dataname{} is challenging, indicating that LLMs, such as OpenAI o1, have insufficient counterexample-driven proof capabilities. Moreover, our exploration into model training reveals that strengthening LLMs' counterexample-driven conceptual reasoning abilities is crucial for improving their overall mathematical capabilities. We believe that our work offers new perspectives on the community of mathematical LLMs.
\end{abstract}

\section{Introduction}
\label{sec:introduction}
Mathematics, as \textcolor{black}{a fundamental aspect of reasoning}, has \textcolor{black}{garnered significant} research interest. \textcolor{black}{Recent studies have demonstrated that Large Language Models (LLMs) exhibit strong mathematical reasoning abilities}~\cite{openai2023gpt4, geminiteam2024gemini, yang2024qwen2, shao2024deepseekmath, ying2024internlm, abel, luo2023wizardmath, yumetamath}. 
\textcolor{black}{Enhancing the mathematical reasoning capabilities of LLMs} has become a \textcolor{black}{prominent} and fundamental topic \textcolor{black}{within} the LLMs research community.


\begin{figure*}
    \centering
\includegraphics[width=0.82\linewidth]{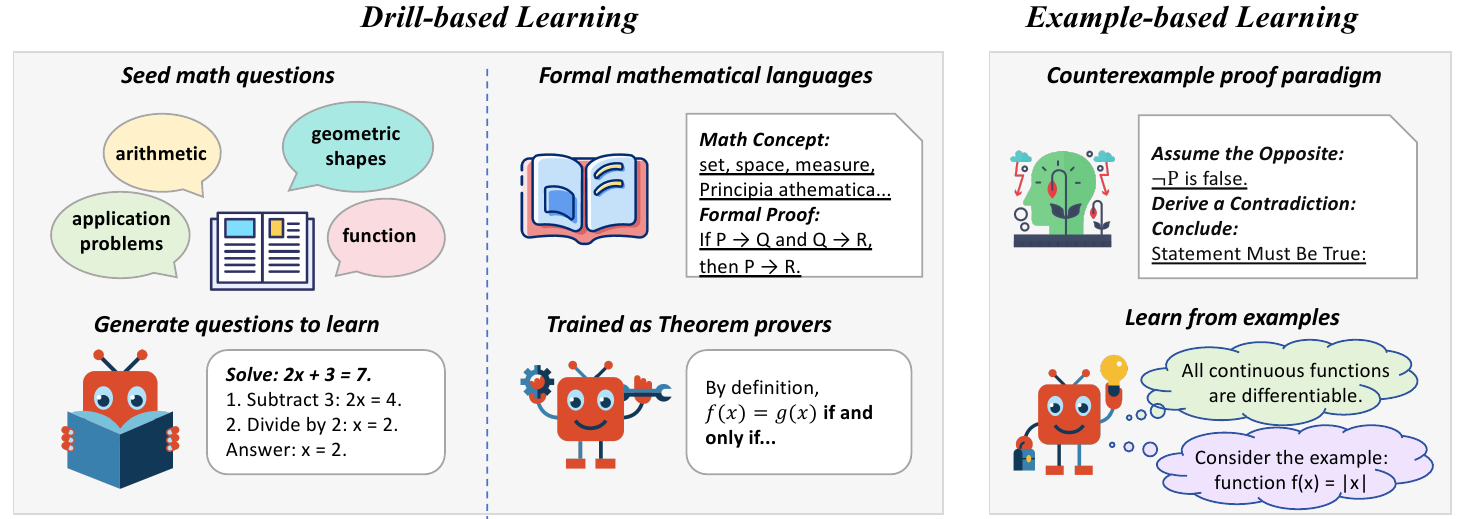}
    \caption{Comparison between drill-based learning and example-based learning. The first two math LLMs fail when confronted with advanced mathematics, and ``Proving by examples'' is a highly creative and concept-intensive mathematical skill.}
    \label{fig:teaser}
\end{figure*}

\textcolor{black}{Currently,} there are two main paradigms for enhancing the mathematical reasoning capabilities of LLMs. \textit{The first \textcolor{black}{involves} synthetic generation based on seed math questions}~\cite{yu2023metamath, li2024common}. 
For example, 
WizardMath~\cite{luo2023wizardmath} introduces a variety of math instructions to generate math questions \textcolor{black}{of} different complexities \textcolor{black}{using} GPT-3.5. 
\textit{The \textcolor{black}{second approach leverages} formal mathematical languages to train LLM-based theorem provers}, such as Lean 4~\cite{Lean4}. For instance, Draft-Sketch-Prove~\cite{jiangdraft}, HunyuanProver~\cite{li2024hunyuanprover}, and Lean-STaR \cite{lin2024lean} interact with formal languages \textcolor{black}{through} informal proofs, automatic formalization, and natural language thoughts for theorem proving. 

The \textcolor{black}{two methods above enable} LLMs to \textcolor{black}{develop} problem-solving \textcolor{black}{skills either by} training \textcolor{black}{on} massive similar problems, or \textcolor{black}{by gaining} proficiency \textcolor{black}{through exposure to} similar proof processes~\cite{mirzadeh2024gsm, yu2024reasonagain}. In both cases, \textcolor{black}{these approaches enhance LLMs’} mathematical reasoning \textcolor{black}{abilities through training, where} proficiency is \textcolor{black}{achieved through familiarity, akin to} ``drill-based'' learning in human mathematics learning.
However, \textcolor{black}{relying solely on} intensive-practice \textcolor{black}{by inundating LLMs} with math problems is neither sufficient nor essential \textcolor{black}{for true} mathematics learning.
\textbf{In other words, drill-based learning \textcolor{black}{alone does not foster} a deep understanding of mathematical concepts \textcolor{black}{in either} humans or LLMs.}


As illustrated in Figure~\ref{fig:teaser}, for human mathematics learning, \textcolor{black}{``example-based'' learning is a more important strategy than drill-based learning}. \textcolor{black}{In particular,} for mathematical proofs, ``proof by counterexamples'' is an indispensable approach.
Inspired by the \textcolor{black}{idea} that counterexample-driven \textcolor{black}{proofs better} reflect \textcolor{black}{a deep} understanding of mathematical concepts, \textcolor{black}{we propose \dataname{}~\footnote{\url{https://github.com/THUKElab/COUNTERMATH}}, a} counterexample-based mathematical reasoning benchmark. \textcolor{black}{\dataname{} is designed to evaluate LLMs’ ability} to distinguish subtle differences between mathematical terms and properties at university-level by providing examples. Specifically, we collect 1,216 statement-rationale pairs from mathematical textbooks, \textcolor{black}{focusing on disproving certain statements under unusual conditions using counterexamples}. \textcolor{black}{In terms of} difficulty, \dataname{} covers advanced mathematical knowledge similar to PutnamBench \cite{tsoukalasputnambench} and Putnam-AXIOM \cite{gulati2024putnamaxiom}, \textcolor{black}{both of which assess} the depth of mathematical understanding in LLMs.

In addition to \textcolor{black}{extensively evaluating} various mainstream mathematics LLMs on \dataname{}, we also develop a framework for automatically acquiring counterexample-based mathematical reasoning data to enable further model training.
Detailed analyses \textcolor{black}{of both} the evaluated LLMs and our trained LLMs reveal that:
\begin{itemize}
    \item The contemporary LLMs including OpenAI o1 \textcolor{black}{exhibit} limited performance in \textcolor{black}{determining} whether \textcolor{black}{a statement in} \dataname{} is true or false, indicating significant room for improvement in \textcolor{black}{higher-level mathematical} conceptual reasoning.  

    \item When analyzing the reasoning process of LLMs, many models still struggle \textcolor{black}{with} example-based reasoning. This demonstrates the limitations of drill-based learning and \textcolor{black}{underscores} the potential value of \dataname{} in advancing mathematical LLMs.  

    \item \textcolor{black}{Lower performance is} observed in topology and real analysis during our fine-grained evaluation, which indicates promising future research directions. \textcolor{black}{Further} studies on mathematical LLMs should explore these underrepresented areas of higher mathematics.  

    \item Our fine-tuned \textcolor{black}{model, trained with} only 1,025 training \textcolor{black}{samples,} demonstrates strong performance on both our benchmark and OOD \textcolor{black}{benchmarks. This confirms} that learning counterexample-based reasoning is not only effective for our task but also \textcolor{black}{has} general significance for improving mathematical reasoning.

\end{itemize}

\begin{figure*}[t]
    \vspace{-4cm}
    \centering
    \includegraphics[width=\linewidth]{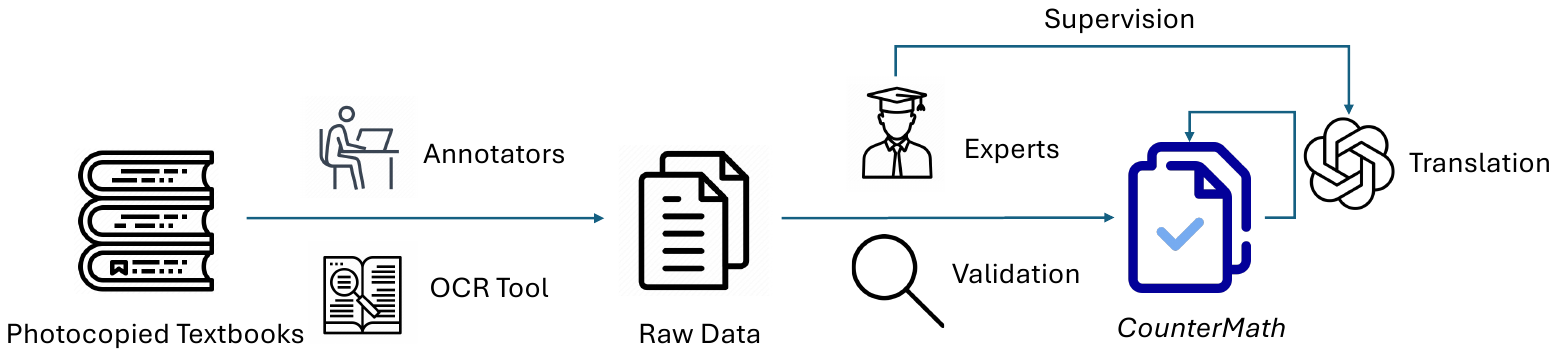}
    \vspace{-5cm}
    \caption{Overview the construction process of \dataname{}. \dataname{} was first extracted from photocopied mathematical textbooks by crowd-sourced labelers with the OCR tool. For the next step, authors with bachelor degrees in applied mathematics as annotation experts would filter and correct improper statement-rationale pairs. Finally, GPT-4o was prompted to translate the validated data into English under experts' supervision.}
    \label{fig:pipeline}
\end{figure*}

\section{Related Work}
\paragraph{Math Benchmarks.} Recently, the number of math-related benchmarks has increased drastically~\cite{amini2019mathqa, yang2019learning, zhengminif2f, hendrycksmath2021, cobbe2021gsm8k, GHOSTS, liu2024mathbench, he2024olympiadbench, lu2024mathvista}. The most influential ones are MATH \cite{hendrycksmath2021} and GSM8K \cite{cobbe2021gsm8k}, which focus on arithmetic reasoning at the high school competition level and grade school level, respectively. Moreover, other benchmarks such as MathBench \cite{liu2024mathbench} and OlympiadBench \cite{he2024olympiadbench} are also blends of problems sets from various competitions and standard examinations, which are used to test human students' abilities of utilizing the math knowledge and certain tricks to solve complex application-based problems. However, mathematicians are more expecting LLMs to help them in literature review, idea generation, proof-checking and collaborative writing as they focus on a broader spectrum of mathematical activities rather \cite{frieder2024largelanguagemodelsmathematicians}. To better accommodate the true need for math research, some formal theorem proving benchmarks like PutnamBench \cite{tsoukalasputnambench}, CoqGym \cite{yang2019learning} and MiniF2F \cite{zhengminif2f} are also proposed recently in a combination of formal mathematical languages compilers (e.g. Coq, Lean), which could be viewed as the important math benchmarks for developing \textit{Mathematics Mechanization} \cite{wen2001mathematics, mathmechanization}.

On the contrary, our benchmark \dataname{} focuses on conceptual reasoning among mathematical concepts and theorems. Specifically, we research certain math reasoning technique: \textit{counterexamples in mathematics}, to check whether the models fully and correctly understand math concepts and theorems, which should be one of atomic abilities for what mathematician are expecting from LLMs compared to independently solving some simple math word problems.

\paragraph{Math Augmented LLMs.} In contrast to general-purpose models such as GPT-4 \cite{openai2023gpt4} and Gemini \cite{geminiteam2024gemini}, several mathematics augmented LLMs have been developed using methods like data augmentation, pretraining, fine-tuning, and reinforcement learning with extensive mathematical corpora. For instance, WizardMath \cite{luo2023wizardmath} employed tailored math prompts to generate seed data and then underwent RLHF and process supervision for training. Abel \cite{abel} utilized supervised fine-tuning with meticulous data processing, referred to as \textit{Parental Oversight}. InternLM2-Math \cite{ying2024internlm} enhanced mathematical reasoning with chain-of-thought \cite{wei2022chain}, code interpreters, and Lean4 translation and theorem proving. NuminaMath \cite{numina_math}, which recently secured first place in the Kaggle AIMO competition\footnote{\url{https://www.kaggle.com/competitions/ai-mathematical-olympiad-prize/leaderboard}}, leveraged \textit{tool-integrated reasoning (TIR)} to generate math questions with fine-grained solutions. Qwen2.5-Math \cite{yang2024qwen2}, initialized with general-purpose Qwen2.5 models, was trained on the undisclosed large-scale and high-quality mathematics-specific corpus. Deepseek-Math \cite{shao2024deepseekmath} focuses on data engineering during pretraining and efficient RL training. 

\paragraph{Conceptual Reasoning.} Conceptual reasoning is an ability to reason in abstract and high-level perspectives \cite{wang2024role,DBLP:conf/coling/HuangMLHZ0Z24, li2024llms,li2025correct}. Recently, there are numerous studies where LLMs are reasoning on abstracted and conceptualized structures by analogy, deduction, induction, etc \cite{saparov2023testing,li2023towards,yasunaga2024large,xu2024let,wang2024hypothesis,cheng2024inductive, zhou2024conceptual}. Specifically in math, conceptual reasoning requires people to reason around math concepts and axioms at the play of math hypothesis, statements and problems \cite{simon2011studying}. An example of this is ConceptMath \cite{wu2024conceptmath}, a math word problem benchmark in elementary school and middle school level, but the reasoning in solving these problems remains superficial as it just requires models to extract the correct variables and do basic arithmetic operations and it is also saturated with GPT models, which diminishes it from showing whether LLMs are truly mastering mathematics.

\section{\dataname}

\subsection{Data Construction}
\label{sec:data_construct}
Our dataset is constructed from a series of math textbooks focusing on counterexamples in different fields such as Algebra \citep{counteralgebra}, Topology \citep{countertopology}, Real Analysis \citep{counterrealanalysis} and Functional Analysis \citep{counterfunctionalanalysis}. \textbf{We have obtained the authors’ consent to use their publications solely for academic research purposes.} As the raw data sources are in Chinese, we also translate our dataset into English, \textcolor{black}{creating a mathematical} conceptual reasoning benchmark based on counterexamples, named as \dataname{}. \textcolor{black}{Each data point includes a} \texttt{statement}, \texttt{rationale}, \texttt{judgement} (i.e., whether the statement is True or False by its rationale), and \texttt{field}.

As illustrated in Figure~\ref{fig:pipeline}, we first recruited several Chinese annotators from \textcolor{black}{specific} vendors to extract statement-rationale pairs from the \textcolor{black}{aforementioned} textbooks \textcolor{black}{using an} OCR tool, which yielded 1,274 statement-rationale pairs. Next, \textcolor{black}{the experts, among the authors, who have the bachelor's degrees in applied mathematics} manually checked \textcolor{black}{all the} data points from the previous stage, \textcolor{black}{annotated each statement as True or False} based on its rationale, and filtered \textcolor{black}{out ambiguous pairs}. This resulted in 1,216 data samples as the final version of our dataset. \textcolor{black}{Additionally, since} the data source \textcolor{black}{was} originally in Chinese, we prompted GPT-4o\footnote{\url{https://openai.com/index/hello-gpt-4o/}} to translate the dataset into English. The authors \textcolor{black}{then further} validated the translated dataset to \textcolor{black}{ensure its} correctness and appropriateness. More details about the annotation process are provided in Appendix~\ref{appendix:data}.

\subsection{Data Analysis}
\label{data:analysis}
\begin{figure}[htbp]
    \centering
    \begin{subfigure}[b]{0.25\textwidth}
        \centering
        \includegraphics[width=\textwidth]{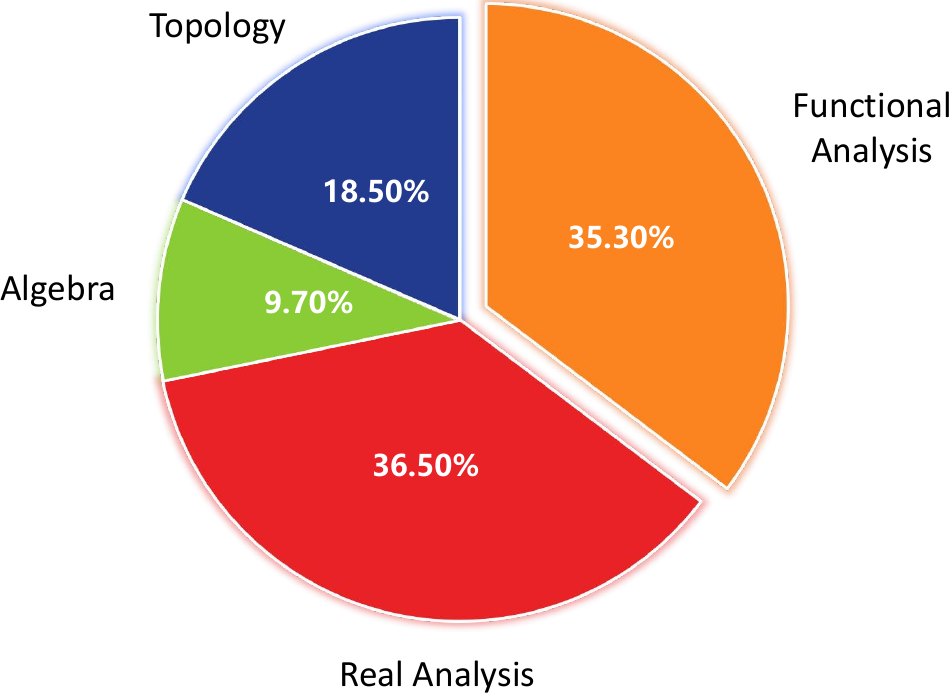}
        \caption{Different fields.}
        \label{fig:data-dist}
    \end{subfigure}
    \hfill
    \begin{subfigure}[b]{0.18\textwidth}
        \centering
        \includegraphics[width=\textwidth]{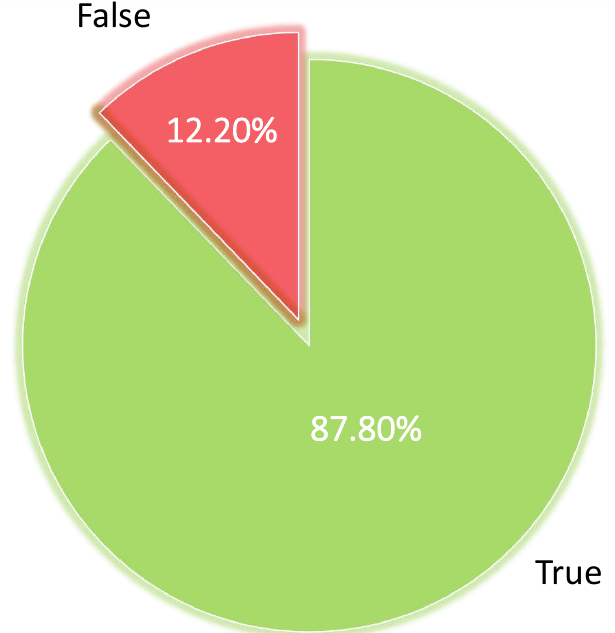}
        \caption{Judgement types.}
        \label{fig:data-dist-judge}
    \end{subfigure}
    \caption{Data Distribution of \dataname{}.}
    \label{fig:combined}
\end{figure}

\paragraph{Overview} \textcolor{black}{Since} the dataset has been constructed from textbooks in four different fields, the distribution of statement-rationale pairs \textcolor{black}{is shown} in Figure~\ref{fig:data-dist}. Moreover, the distribution of judgements \textcolor{black}{is presented} in Figure~\ref{fig:data-dist-judge}. We observe that most statements are labeled as True. This \textcolor{black}{may} be because the data is sourced from mathematical \textcolor{black}{textbooks, where} most statements are \textcolor{black}{phrased correctly to avoid misleading} readers, especially novices in mathematics. In general, as shown in Figure~\ref{fig:annt_example}, the \textcolor{black}{statements often involve} several college-level mathematical concepts or properties, focusing on nuanced understandings \textcolor{black}{of} mathematics. In real-world applications, due to the concise formulations, these statements are \textcolor{black}{frequently} used as interview questions in math-related graduate \textcolor{black}{programs. This also contributes} to \textbf{the diversity of mathematical testbeds for contemporary \textcolor{black}{LLMs, fostering research} in conceptual mathematical reasoning} \cite{calculusbook}.

\paragraph{Data Validation} As mentioned in Section~\ref{sec:data_construct}, the retention rate between the two annotation stages is 95.4\%, demonstrating that most high-quality statement-rationale pairs \textcolor{black}{were} \textcolor{black}{extracted} from the textbooks. The filtered statement-rationale pairs often suffer \textcolor{black}{from issues such as} \textit{irrelevance between statement and rationale}, \textit{\textcolor{black}{excessive} typos}, and \textit{trivial \textcolor{black}{rationales,} such \textcolor{black}{as simply} quoting conclusions from other statement-rationale pairs or even papers}.

\section{Benchmark Settings}
\paragraph{Baselines} We are testing following large language models with \dataname{}. For \textit{open-weight models}, we use \textbf{Deepseek-Math-7B-RL} \cite{shao2024deepseekmath}, \textbf{Eurus-2-7B-PRIME} \cite{cui2024process}, \textbf{Qwen2.5-Math-7B/72B-Instruct} \cite{yang2024qwen2}, \textbf{NuminaMath-7B-TIR} \cite{numina_math}, \textbf{InternLM2-Plus-7B/20B/Mixtral8x22B} \cite{ying2024internlm}, \textbf{Abel-7B/13B/70B} \cite{abel}, \textbf{WizardMath-7B/70B} \cite{luo2023wizardmath}, \textbf{Mathstral-7B}\footnote{\url{https://mistral.ai/news/mathstral/}}, \textbf{MetaMath-Mistral-7B} \cite{yu2023metamath}, \textbf{Xwin-Math-7B/13B/70B} \cite{li2024common}, \textbf{Rho-math-7b-interpreter} \cite{lin2024rho}, \textbf{MAmmoTH2-7B/8x7B-Plus} \cite{yue2024mammoth2}, \textbf{QwQ-32B-Preview}\footnote{\url{https://qwenlm.github.io/blog/qwq-32b-preview/}}. For \textit{proprietary models}, we use \textbf{GPT-4o}, \textbf{OpenAI o1}\footnote{\url{https://cdn.openai.com/o1-system-card-20241205.pdf}}, \textbf{Qwen-Max}\footnote{\url{https://www.alibabacloud.com/help/en/model-studio/developer-reference/what-is-qwen-llm}}, \textbf{Deepseek-R1}\footnote{\url{https://api-docs.deepseek.com/news/news250120}}, \textbf{Claude3.7-sonnet}\footnote{\url{https://www.anthropic.com/claude/sonnet}}, and \textbf{Gemini2.5-pro}\footnote{\url{https://deepmind.google/technologies/gemini/pro/}}. The selection of baseline models tries to cover as many representative models as possible, which considers various perspectives, including data processing, training paradigms, base models, and developers (i.e., academia or industry). The detailed summary of open-weight baseline models is in Appendix~\ref{appendix:experiments}.

\paragraph{Prompts} \textcolor{black}{To} maximize the performance of LLMs on \dataname{}, we use the default \textbf{CoT} prompts for each LLM, which are \textcolor{black}{typically} mentioned in their Huggingface model cards. These prompts include completion prompts, Alpaca-like prompts, and chat-template-based prompts. \textcolor{black}{Additionally,} we adopt a \textbf{Hint} \textcolor{black}{prompt, designed to encourage LLMs} to provide examples when attempting to solve problems. \textcolor{black}{A summary of our used prompts is provided} in Appendix~\ref{appendix:experiments}. 
\textcolor{black}{Note that it is observed that using In-Context Learning (ICL) prompts \textcolor{black}{do} not significantly affect the performance of LLMs on \dataname{}. We \textcolor{black}{believe this is due to the nature of} counterexample-driven conceptual reasoning, where LLMs \textcolor{black}{struggle} to learn the ability to \textcolor{black}{provide examples from} a small number of demonstrations. In particular, \textcolor{black}{mathematical subfields exhibit substantial variation in concepts and terminology}. Therefore, we do not use \textcolor{black}{ICL prompts} in our experiments.}

\begin{figure*}[t]
    \centering
    \includegraphics[width=0.95\linewidth]{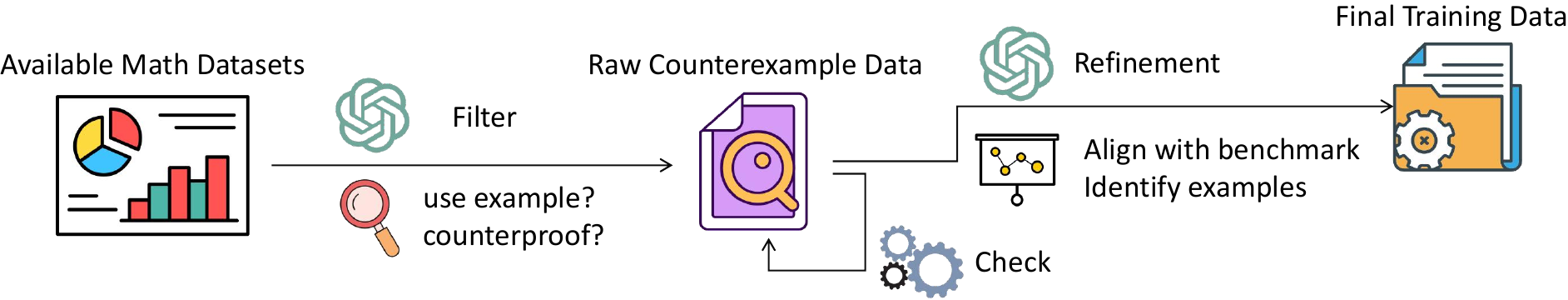}
    \caption{The overview of our training data engineering framework.}
    \label{fig:traingdata}
\end{figure*}

\paragraph{Evaluation Metrics} Our evaluation metrics are two-fold. For efficiency, we use lexical matching such as \textit{F1} to match the judgements of the statements. The reason of not using \textit{accuracy} is the imbalance of data distributions mentioned in Section~\ref{data:analysis}. 
To assess whether the model has acquired the capability of solving mathematical problems through exemplification, we conducted a targeted evaluation in this section. We designed a systematic evaluation framework leveraging GPT-4o as an automated judge to perform the following tasks: \textbf{Example Extraction} automatically identifies and extracts instances where the model explicitly \textcolor{black}{uses} exemplification (i.e., generating or referencing specific counterexamples to justify the statement) during its reasoning process. \textbf{Alignment Assessment} evaluates whether each extracted example aligns with the reasoning approach of a predefined Reference Example in terms of logical structure, problem decomposition, and goal relevance. The evaluation prompts are \textcolor{black}{detailed} in Appendix~\ref{appendix:experiments}. \textcolor{black}{We also} designed three evaluation metrics to assess the models' ability to solve problems by providing examples:
\begin{itemize}
    \item Proportion of \textbf{Examples}: This metric calculates the proportion of problem-solving cases in which the model employs examples as part of its solution.
    \item \textbf{Strict} Align: This measures the percentage of the model's provided examples that are fully consistent with the reference.
    \item \textbf{Loose} Align: This evaluates the proportion of instances where at least one example provided by the model aligns with the reference.
\end{itemize}
To further validate the reliability of the proposed LLM-based metrics, we conducted a comprehensive human evaluation. This evaluation focused primarily on two key aspects: (1) the accuracy of example extraction, and (2) the consistency of the evaluation results with human judgement. We \textcolor{black}{selected} 100 sample cases and manually \textcolor{black}{reviewed} them. \textcolor{black}{First}, the accuracy of the example extraction process, as performed by the model, was found to \textcolor{black}{be} 97\%, demonstrating the robustness of our automatic extraction mechanism. \textcolor{black}{Second,} the evaluation of the extracted examples—assessed for alignment with reference examples—achieved an accuracy of 93.5\% when compared to human judgements. These results indicate a strong alignment between the model's automatic evaluation process and human evaluators, confirming the reliability and validity of our methodology for assessing the use of exemplification in mathematical problem-solving.

\section{Conceptual Finetuning}
\subsection{Training Data Engineering Framework}
\paragraph{Filter-based Data Collection}
To validate our \textcolor{black}{approach}, we conduct supervised fine-tuning to enhance the model's ability to provide examples for conceptual reasoning and incentivize its higher-level mathematical understanding. 
As shown in Figure \ref{fig:traingdata}, we propose an automatic training data engineering framework to obtain training data.
\textcolor{black}{Since most} LLMs currently \textcolor{black}{cannot provide} satisfactory mathematical proofs based on examples, for data collection, we filter data from existing human-written datasets \textcolor{black}{rather than} directly generating counterexample data.
Specifically, \textcolor{black}{we apply a} data filtering and refinement strategy based on strictly labeled datasets of propositional statements and proofs. We collect several high-quality mathematical proof datasets, such as ProofNet \cite{proofnet} and NaturalProof \cite{naturalproof}, \textcolor{black}{ensuring} that there is no overlap between these datasets and our \dataname{}. We use GPT-4o to filter data specifically involving proofs using counterexamples. Our designed data filtering prompt is presented in Appendix~\ref{appendix:training_data}. Inspired by AutoRace \cite{hao2024llm} \textcolor{black}{, which applies} LLM as a judge, we design several criteria to evaluate whether the data utilize counter-reasoning, counter-examples for proof, or special examples. To maximize data retention, we employ a loose filtering criterion, retaining data if at least one criterion \textcolor{black}{is} met. Subsequently, we double-check the filtered data by assessing the completeness of statements, and the rigor and validity of proofs, while discarding incomplete or inconsistent entries. We ultimately obtain 1,025 filtered samples from an initial pool of over 30,000 data.

\begin{table*}[t]
\centering
\caption{Main evaluation results of various mainstream mathematical LLMs with the default CoT prompts on \dataname{}. The \textbf{Examples, Strict, and Loose} represent the three of our designed example-related evaluation metrics.}
\vspace{3mm}
\resizebox{0.85\linewidth}{!}{
\begin{tabular}{l|l|cccc}
\toprule[1.3pt]
\multicolumn{2}{c|}{\multirow{2}{*}{\textbf{Models}}}                                               & \textbf{Judgement}    & \multicolumn{3}{c}{\textbf{Rationale Reasoning}}   \\
\multicolumn{2}{l|}{}                                                                      & \textbf{F1   (macro)} & \textbf{Examples}   (\%) & \textbf{Strict} (\%) & \textbf{Loose} (\%) \\ \midrule[1.3pt]
\multicolumn{6}{c}{\textbf{\textit{Open source models}}}                                                                                                               \\ \midrule
\multirow{12}{*}{size = 7B}                     & Deepseek-Math-7B-rl                     & 32.2         & 65.9           & 18.9        & 20.6       \\
                                                & Eurus-2-7B-PRIME                 & 37.5         & 64.8           & 28.5        & 32.0       \\
                                                & NuminaMath-7B-TIR                 & 30.4         & 54.1           & 13.0        & 13.7       \\
                                                & InternLM2-Math-Plus-7B                  & 33.9         & 36.6           & 9.0         & 9.5        \\
                                                & Abel-7B-002                             & 34.4         & 66.1           & 16.0        & 17.9       \\
                                                & WizardMath-7B-v1.1                & 27.9         & 43.2           & 6.4         & 7.2        \\
                                                & Mathstral-7B-v0.1                 & 28.2         & 38.9           & 7.5         & 7.9        \\
                                                & MetaMath-Mistral-7B                     & 31.0         & 26.5           & 0.4         & 0.7        \\
                                                & Xwin-Math-7B-V1.0                 & 28.1         & 31.3           & 1.2         & 1.7        \\
                                                & rho-math-7b-interpreter-v0.1      & 22.3         & 18.3           & 1.9         & 2.1        \\
                                                & MAmmoTH2-7B-Plus                  & 32.3         & 54.2           & 10.7        & 12.1       \\
                                                & Qwen2.5-Math-7B-Instruct          & \textbf{38.3}         & \textbf{74.2}           & \textbf{30.2}        & \textbf{33.2}       \\ \midrule
\multirow{5}{*}{7B\textless size \textless 70B} & Abel-13B-001                      & 22.4         & 24.4           & 0.8         & 0.8        \\
                                                & Xwin-Math-13B-V1.0                & 30.2         & 31.3           & 1.2         & 1.7        \\
                                                & InternLM2-Math-Plus-20B                 & 18.4         & 28.8           & 8.4         & 9.5        \\
                                                & MAmmoTH2-8x7B-Plus                & 28.8         & 51.4           & 14.1        & 15.5       \\
                                                & QwQ-32B-Preview                   & \textbf{39.9}         & \textbf{70.0}           & \textbf{38.6}        & \textbf{43.8}       \\ \midrule
\multirow{5}{*}{size \textgreater{}=70B}        & InternLM2-Math-Plus-Mixtral8x22B  & 37.3         & 63.2           & 21.5        & 23.1       \\
                                                & Xwin-Math-70B-V1.0                & 25.5         & 25.2           & 1.4         & 1.7        \\
                                                & Abel-70B-001                      & 31.0         & 48.4           & 5.3         & 6.1        \\
                                                & WizardMath-70B-v1.0               & 24.2         & 52.9           & 6.3         & 7.4        \\ 
                                                & Qwen2.5-Math-72B-Instruct       & \textbf{41.8}         & \textbf{76.6}           & \textbf{38.9}        & \textbf{41.6}       \\ \midrule
\multicolumn{6}{c}{\textbf{\textit{Commercial models}}}                                                                                                                \\ \midrule
\multicolumn{2}{c|}{GPT-4o}                                                                & 59.0         & 44.7           & 19.7        & 21.3       \\
\multicolumn{2}{c|}{OpenAI o1-preview}                                                             & 60.1         & 55.8           & 39.8        & 40.9       \\
\multicolumn{2}{c|}{Qwen-max}                                                              & 58.9         & 61.8           & 30.4        & 33.9      \\ 
\multicolumn{2}{c|}{Deepseek-R1}                                                              & \textbf{80.7}         & 86.8           & 54.2        & 65.3      \\
\multicolumn{2}{c|}{Claude3.7-sonnet}                                                                & 64.8         & 78.0           & 45.0        & 52.5       \\
\multicolumn{2}{c|}{Gemini2.5-pro}                                                                & 77.0         & \textbf{90.8}           & \textbf{65.1}        & \textbf{75.7}       \\
\bottomrule[1.3pt]
\end{tabular}}
\label{tab:main}
\end{table*}

\paragraph{Training Data Refinement}
We further refine the rationale of the collected SFT data to improve the model's ability to effectively provide examples during training. Compared with our \dataname{}, we observe that the SFT data \textcolor{black}{feature} longer proof processes and \textcolor{black}{lack} explicit illustrations of the provided examples. To address this, we employ GPT-4o to refine the rationales to better align with the feature distribution of \dataname{}. Specifically, we randomly select one example from each of the four fields in our \dataname{} as a reference, and provide three manually rewritten before-and-after comparisons to guide the model. Our designed data refinement prompt is presented in Appendix~\ref{appendix:training_data}. We emphasize that while reducing proof length, the rewriting process should preserve the original reasoning structure to maintain the integrity of the data. This refinement \textcolor{black}{ensures} that the SFT \textcolor{black}{data closely aligns with the characteristics of \dataname{}} and explicitly incorporates example usage to facilitate training.

\subsection{Training and Evaluation Details}
For model training, we select Qwen-2.5-Math-7B-Instruct, an open-source model known for its strong mathematical reasoning capabilities and general applicability. we fine-tune Qwen2.5-Math-Instruct-7B using supervised LoRA training on 2$\times$L20 48GB GPUs, with a learning rate of 1.0e-5. After training, we evaluate the model on both our \dataname{} and several out-of-distribution (OOD) benchmarks, such as MATH\cite {hendrycksmath2021} and GSM8K \cite{cobbe2021gsm8k}, OlympaidBench \cite{he2024olympiadbench}, MMLU \cite{wang2025mmlu}, AIME 2024. We aim to demonstrate not only improved performance on the \dataname{} but also enhanced generalization to OOD benchmarks, thereby validating the effectiveness of our insight in improving the model's overall conceptual reasoning capabilities by examples.

\section{Analysis and Discussions}
\subsection{Evaluation Results without Finetuning}

We selected a range of advanced mathematical LLMs with varying parameter sizes to evaluate their conceptual reasoning abilities on our benchmark. Table \ref{tab:main} summarizes their performance across various metrics. From the results, we derive the following findings and insights:

\begin{figure}[t]
    \centering
    \includegraphics[width=0.8\linewidth]{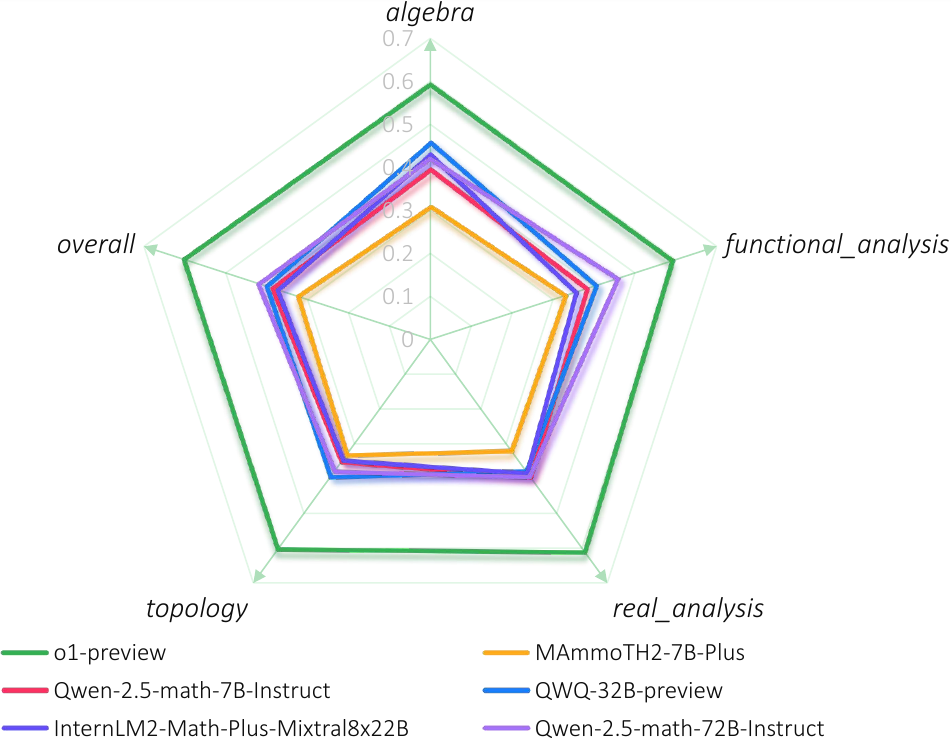}
    \caption{Fine-grained evaluation results of different fields in \dataname{}. }
    \label{fig:field}
\end{figure}

\paragraph{Judgement Performance}
The performance on the automatic evaluation metric F1 reflects the models' fundamental conceptual reasoning \textcolor{black}{abilities, specifically their} capacity to correctly determine the truth or falsehood of a given statement. While open-source models exhibit some performance, their overall performance is relatively low around 30. Even the advanced Qwen-2.5-Math-72B-Instruct achieves only 41.8, falling behind commercial models. Notably, the math-reasoning-specific Deepseek-R1 model outperforms others, achieving an F1 score of 80.7. However, compared to benchmarks focused on elementary and high school mathematics, \textcolor{black}{the overall} performance of commercial models is relatively lower. This disparity underscores the challenges posed by our higher-level mathematics benchmark, aligning with our objective of exploring deeper conceptual reasoning. 
\textcolor{black}{When \textcolor{black}{comparing} the performance of models of different sizes, we find that some small 7B models outperform the larger 72B model. For example, Deepseek-Math-7B-rl \textcolor{black}{outperforms} WizardMath-70B-v1.0. We believe this phenomenon is related to the training methods behind \textcolor{black}{these} models. As explained in Section~\ref{sec:introduction}, drill-based training \textcolor{black}{in} WizardMath-70B-v1.0 is not very effective in improving the ability to understand deep and complex mathematical \textcolor{black}{concepts. In contrast, the superior} performance of Deepseek-Math-7B-rl \textcolor{black}{suggests} that reinforcement learning may \textcolor{black}{play a key role in enhancing} counterexample-driven conceptual reasoning.}

\paragraph{Conceptual Reasoning Ability with Examples}
Beyond assessing binary classification F1 in judgements, we also evaluate the reasoning process of the models. We hope that \textcolor{black}{LLMs provide} logical proofs or counterexamples rather than relying on random guessing or copying statements. \textcolor{black}{The Qwen-series} models demonstrate superior higher mathematics reasoning, accurately identifying statements suited for counterexample reasoning and surpassing even the commercial o1 model. However, many open-source models fail to generate meaningful examples. For instance, MetaMath and rho produce counterexamples in only 26.5\% and 18.3\% of cases, respectively, with consistency rates as low as 0.4\% and 1.9\%. This limitation likely originates from the ``Practice lots of math problems'' training strategies employed, where most data are derived from elementary and high school mathematics. Such training leaves these models ill-equipped to handle the abstraction and conceptual reasoning demands of higher-level mathematics.

\paragraph{Fine-grained Analysis Across Fields}
We further conduct a fine-grained analysis of four mathematical fields in \dataname{}, evaluating six advanced models of varying sizes. As shown in Figure \ref{fig:field}, the o1 model consistently outperforms others across all fields, with relatively balanced performance in each. Among open-source models, we observe stronger performance in algebra and functional analysis but weaker results in topology and real analysis. This suggests that topology and real analysis pose greater challenges for LLMs, possibly due to their lower occurrence in training data. These findings highlight the need for future research to focus on developing LLMs capable of handling these underexplored fields of mathematical reasoning.

\begin{table*}[h]
\centering
\caption{The evaluation results on our \dataname{}.}
\vspace{3mm}
\begin{tabular}{l|c|ccc}
\toprule[1.3pt]
\textbf{Models}                                      & \textbf{F1 (macro)} & \textbf{Examples(\%)} & \textbf{Strict(\%)} & \textbf{Loose(\%)} \\ \midrule[1.3pt]
\multicolumn{5}{l}{\textit{\textbf{Base models}}}                                                               \\ \midrule
Qwen2.5-Math-7B-Instruct                   & 38.3      & 74.2        & 30.2      & 33.2     \\
Qwen2.5-Math-7B-Instruct + Hint prompt     & 39.4     & 79.0          & \textbf{33.1}       & \textbf{36.4}      \\ \midrule
\multicolumn{5}{l}{\textit{\textbf{Our training model}}}                                                        \\ \midrule
Qwen2.5-Math-7B-Instruct-SFT               & 39.7     & 75.2       & 31.4       & 34.7     \\
Qwen2.5-Math-7B-Instruct-SFT + Hint prompt & \textbf{41.1}     & \textbf{79.4}       & 31.1      & 34.7     \\ \bottomrule[1.3pt]
\end{tabular}\label{tab:ours}
\end{table*}

\begin{table*}[h]
\centering
\caption{The Out-of-distribution Evaluation Results.}
\vspace{3mm}
\resizebox{0.9\linewidth}{!}{
\begin{tabular}{l|ccccc}
\toprule[1.3pt]
\textbf{Models}                                                                              & \textbf{GSM8K} & \textbf{MATH} & \textbf{Olympaidbench-Math} & \textbf{MMLU-college-Math} & \textbf{AIME 2024}\\ 
\midrule
Qwen2.5-math-7B-Instruct                                                            & 95.1  & 80.5  & 41.6 & 74.0 & 20.0 \\
Qwen2.5-math-72B-Instruct                                                           & 95.4  & 84.9 & \textbf{49.2}   & 78.3   & 26.7    \\ \midrule
\begin{tabular}[c]{@{}l@{}}Qwen2.5-math-7B-Instruct\\ +Countermath-SFT\end{tabular}  & \textbf{95.6}  & \textbf{87.9}    & 46.4  & \textbf{80.0}  & \textbf{30.0}              \\  \bottomrule[1.3pt]
\end{tabular}}\label{tab:OOD}
\end{table*}

\subsection{Results with Finetuning}


\paragraph{Evaluation on Our Benchmark}
We evaluate the approach on our benchmark to verify whether counterexample reasoning effectively enhances a model's conceptual reasoning capabilities. In addition to the base model, we include comparisons with the hint prompt, where explicit hints encouraging LLM to reason by example were provided. For a fair comparison, we use identical prompts to generate outputs during the experiments.
The results indicate that, with just 1,025 training samples, the fine-tuned model outperforms base models in F1 score. Furthermore, the trained model demonstrates superior example-based conceptual reasoning abilities, showing improvements in both the quantity and quality of examples compared to base models. On the other hand, although the constructed training data are refined to align closely with our \dataname{} distribution, some discrepancies remain. Since our exploration involves only the simple SFT strategy on a limited dataset, the model performs slightly worse on some metrics, which can be considered an acceptable limitation.

\paragraph{Evaluation on Out-of-Distribution Benchmarks}
To assess the generalizability of the fine-tuned model, we further evaluate its performance on out-of-distribution (OOD) benchmarks. This \textcolor{black}{aims to verify} whether the model's counterexample reasoning capability, \textcolor{black}{which is valid} on our \dataname{}, \textcolor{black}{can} transfer to other benchmarks and deliver broader performance improvements. Using identical prompts and configurations for fairness, we compare the base models and fine-tuned models on OOD benchmarks.
The results reveal that the fine-tuned model outperforms the base models on both benchmarks, even surpassing larger models like the 72B-parameter model. This aligns well with our hypothesis: equipping the model with counterexample reasoning ability can enhance conceptual reasoning across the general mathematical domain.

\begin{figure}[ht]
    \centering
    \includegraphics[width=1.00\linewidth]{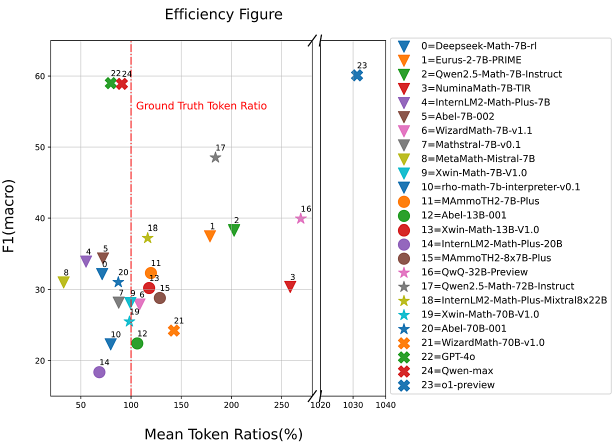}
    \caption{The relationship between Mean Token Ratios (\%) and F1 (macro) scores for various models. The red dashed line represents the Ground Truth Token Ratio (100\%), serving as an efficiency benchmark. Models closer to this line are more token-efficient, while those farther to the right consume significantly more tokens.}
    \label{fig:token used}
\end{figure}

\subsection{Used Tokens Analysis}
The relationship between model performance and token usage efficiency is a critical factor in understanding the trade-offs inherent in model design. As depicted in Figure~\ref{fig:token used}, we analyzed the connection between the mean token ratios (\%) \textcolor{black}{, which represent} the efficiency of token usage relative to ground \textcolor{black}{truth, and} the F1 (macro) score, which reflects the predictive performance of the model. \textcolor{black}{The Mean Token Ratio} (\%) is calculated by dividing the number of tokens actually used by the model during inference by the number of tokens in the ground truth answer. Reasoning models such as "23=o1-preview" and "16=QwQ-32B-Preview" utilize a significantly large number of tokens during inference, but this extensive token consumption does not lead to a corresponding improvement in F1 (macro) scores. This suggests that the benchmark task \textcolor{black}{is highly difficult}, where simply increasing the length or detail of token reasoning does not necessarily enhance performance. Models like "24=Qwen-max" and "22=GPT-4o" demonstrate a commendable balance between token usage and performance. These models achieve relatively high F1 (macro) scores while maintaining token consumption close to the ground truth token ratio. This indicates their ability to perform accurate reasoning efficiently, without overly relying on additional token usage.


\begin{figure*}
    \centering
    \includegraphics[width=0.9\linewidth]{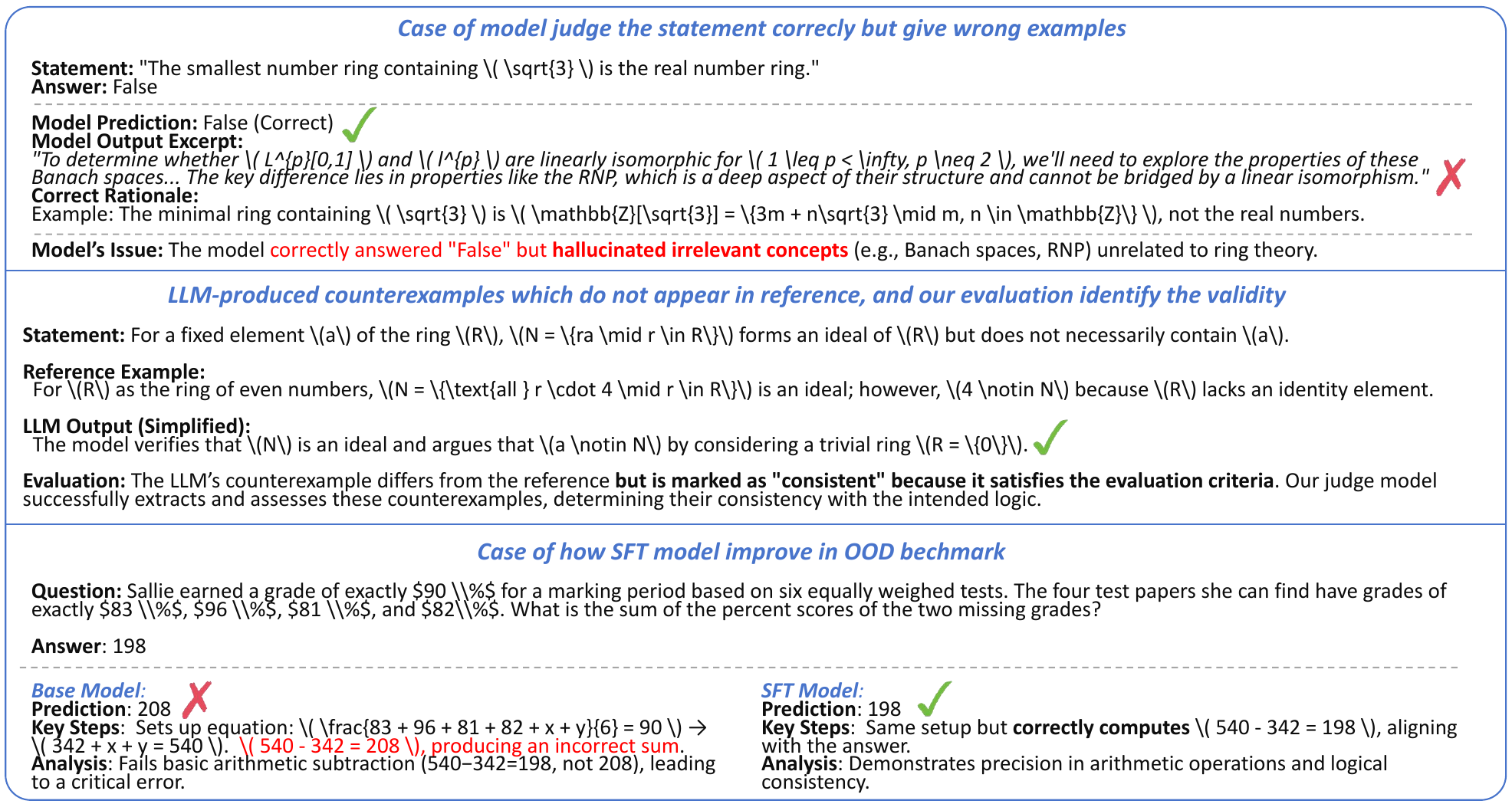}
    \caption{Our case studies of the performance of LLM include: (1) correct judgment with incorrect examples, (2) validation of novel but effective counterexamples, and (3) performance improvement on MATH dataset after counterexample training}
    \label{fig:case}
\end{figure*}
\subsection{Case Study}
In our case study, we observe that while the LLM correctly judges mathematical statements, it sometimes provides incorrect examples as shown in Figure \ref{fig:case}, Case 1. In addition, we verify the validity of our evaluation metric of EXAMPLE. The LLM generates valid counterexamples that differs from reference ones, yet are correctly validated by our judge model, highlighting the effectiveness of our evaluation method. Finally, we compare the improvement of MATH benchmark between base and SFT model, demonstrating the benefit of incorporating counterexample traning to enhance mathematical reasoning capabilities.

\section{Conclusion}
In this work, we address the limitations of drill-based learning in mathematical LLMs by introducing \dataname{}, a counterexample-based reasoning benchmark. Unlike existing datasets, \dataname{} evaluates models on their ability to distinguish nuanced mathematical concepts through example-driven reasoning.
Our key contributions include constructing a high-quality dataset with 1,216 university-level counterexample-based proofs, benchmarking state-of-the-art mathematical LLMs to reveal their conceptual reasoning gaps, and developing an automated framework for counterexample data generation and training. Experimental results show that current LLMs struggle with counterexample-based reasoning, particularly in topology and real analysis, highlighting areas for future research. Furthermore, our fine-tuned model, trained on only 1,025 examples, significantly outperforms baseline models, demonstrating the effectiveness and generalizability of counterexample-driven learning in mathematical reasoning.

\section*{Acknowledgements}
This research is supported by National Natural Science Foundation of China (Grant No. 62276154), Research Center for Computer Network (Shenzhen) Ministry of Education, the Natural Science Foundation of Guangdong Province (Grant No. 2023A1515012914 and 440300241033100801770), Basic Research Fund of Shenzhen City (Grant No. JCYJ20210324120012033, JCYJ20240813112009013 and GJHZ20240218113603006), the Major Key Project of PCL (NO. PCL2024A08), the Key-Area Research and Development Program of Guangdong Province No.2024B1111100001. This work is supported in part by NSF under grants III-2106758, and POSE-2346158.


\section*{Impact Statement}
This paper introduces a conceptual mathematical benchmark aimed at advancing research on Large Language Models (LLMs) in the realm of genuine mathematical reasoning. The dataset utilized in this work is derived from publications with copyrights reserved and authors' permission only for academic research purposes. It is important to acknowledge that our experiments and evaluations rely heavily on LLMs, but this study does not fully explore or mitigate potential biases inherent in their outputs. Addressing these biases and ensuring model alignment with social values remain critical challenges. This underscores the importance of conducting comprehensive evaluations that consider diverse dimensions of human society and their implications.





\bibliography{example_paper}
\bibliographystyle{icml2025}

\newpage
\appendix
\onecolumn
\section{Details for Data Curation}
\label{appendix:data}
\paragraph{Annotation Cost}In the first-stage annotation, there were three annotators recruited who have at least bachelor's degrees in engineering or science-related majors. On average, each of them has annotated about 416 statement-rationale pairs from the given textbooks. They were working through the vendor's provided platform with the provided OCR tool and annotation examples. The cost for the first-stage crowd-sourced annotation was about \$1306.

\begin{figure}[h]
    \centering
    \includegraphics[width=0.9\linewidth]{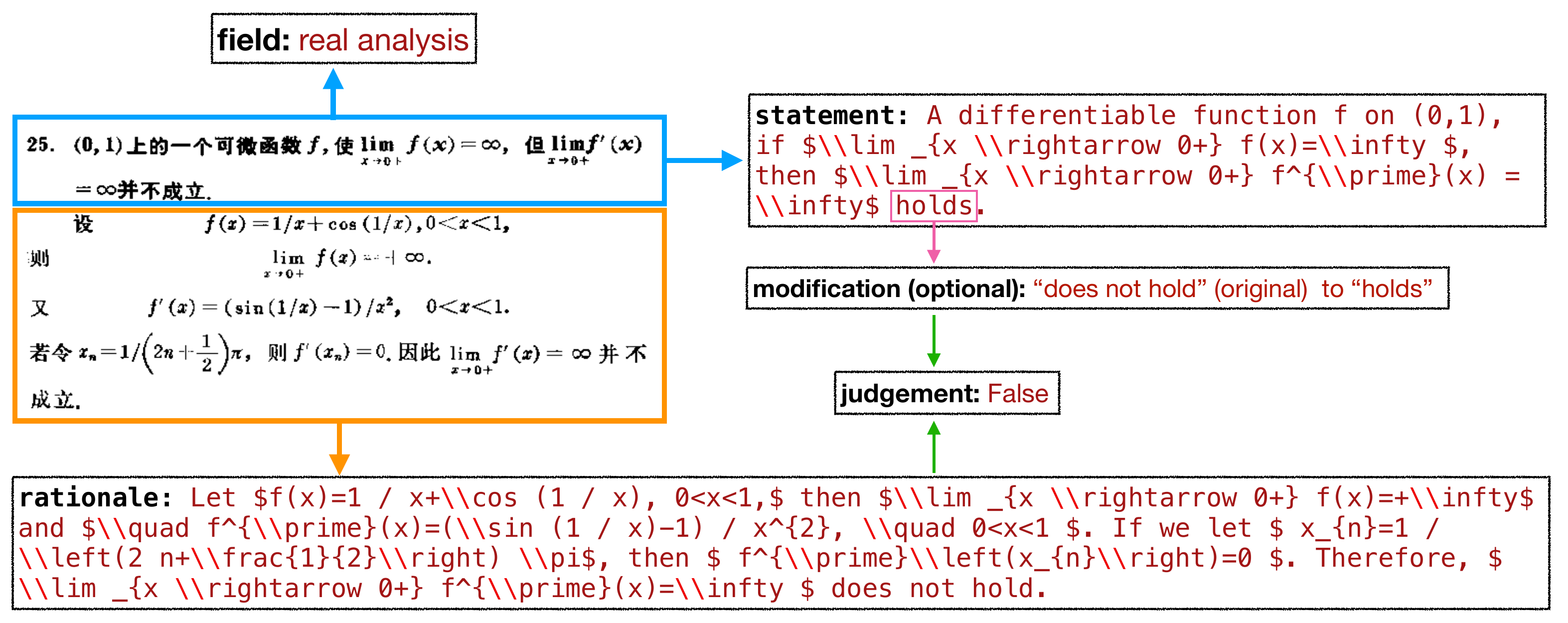}
    \caption{An annotation example from \textit{Counterexamples in Real Analysis.} \cite{counterrealanalysis}.}
    \label{fig:annt_example}
\end{figure}

\paragraph{Annotation Process} The annotation process consists of two stages. In the first stage, the recruited annotator were asked to annotate \texttt{statement}, \texttt{rationale} (i.e. the response for supporting the statement, typically examples or counterexamples), \texttt{field} (i.e. algebra, topology, real analysis or functional analysis), and \texttt{txt} (i.e. mappings to original annotations for the validation use). During annotation, they were provided with several annotation examples created by authors, demonstrating the annotation targets. Moreover, we also asked annotators to focus on statements related to \textit{proving or disproving} and \textit{existence of certain mathematical objects} and ignore those statements without clear or complete answers such as definitions for certain advanced and complicated concepts. For the second stage, authors with at least bachelor's degrees in applied mathematics were checking the LaTeX formats and typos in statements and rationales, keeping the rationales that correct and clearly support or deny the statements with examples or counterexamples. Besides, we also added another element in annotated data points, \texttt{judgement}, to show whether the statement is true or false by its rationale. One full annotation example is shown in Figure~\ref{fig:annt_example}. \textbf{It should be noted that we have modified some statements (less than 5\%), which are easily revised to be the reverse, to make them False for diversity because nearly all statements are stated as True in the original textbooks.} In all, we tried to make sure that all statements and rationales were concise and related.

\section{Details for Experimental Settings}
\label{appendix:experiments}
\begin{table}[h]
\centering
\begin{tabularx}{\linewidth}{@{}cccXl@{}}
\toprule
\multicolumn{1}{c}{\textbf{Models}} &
  \multicolumn{1}{c}{\textbf{Scale}} &
  \multicolumn{1}{c}{\textbf{Base Models}} &
  \multicolumn{1}{c}{\textbf{Training Data}} &
  \multicolumn{1}{c}{\textbf{Training Paradigms}} \\ \midrule
Deepseek-Math-RL &
  7B &
  Deepseek-Math-Base &
  Math corpus from Common Crawl, CoT/PoT/TIR mathematical instruction data, CoT data related to MATH and GSM8K &
  \textit{CP}, \textit{SFT}, \textit{GRPO} \\ \midrule
Qwen2.5-Math-Instruct         & 7B/72B              & Qwen2.5-Math-Base  & Math corpus including Common Crawl and synthetic data, synthetic CoT/TIR math data & \textit{CP}, \textit{SFT}, \textit{GRPO} \\ \midrule
InternLM2-Math-Plus  & 7B/20B/8x22B & InterLM2-Base/Mixtral      & Math corpus from Common Crawl, in-house data and synthetic data, Open math-related instruction data & \textit{CP}, \textit{SFT} \\ \midrule
WizardMath           & 7B/70B              & Mistral/Llama2     & Data augmentation based on MATH and GSM8K & \textit{SFT}, \textit{PPO} \\ \midrule
Abel                 & 7B/13B/70B          & Llama-2            & Data augmentation by Parental Oversight & \textit{SFT} \\ \midrule
Xwin-Math            & 7B/13B/70B          & Llama-2            & Data augmentation based on MATH and GSM8K & \textit{SFT} \\ \midrule
MAmmoTH2-Plus        & 7B/8x7B             & Mistral/Mixtral    & WebInstruct + additional instruction tuning datasets, including Math-Plus, Code-Feedback, etc. & \textit{SFT} \\ \midrule
Eurus-2-PRIME         & 7B                  & Qwen2.5-Math-Base  & Reasoning-related instruction data from open datasets, Curated RL data from NuminaMath, Codeforces, etc. & \textit{SFT}, \textit{PRIME} \\ \midrule
Numina-Math-TIR      & 7B                  & Deepseek-Math-Base & NuminaMath & \textit{SFT} \\ \midrule
Mathstral            & 7B                  & Mistral            & NuminaMath & \textit{SFT} \\ \midrule
MetaMath-Mistral     & 7B                  & Mistral            & Data augmentation based on MATH and GSM8K & \textit{SFT} \\ \midrule
Rho-math-interpreter & 7B                  & Mistral            & OpenWebMath\& general corpora, ToRA(TIR data) & \textit{SLM}, \textit{SFT} \\ \midrule
QwQ-Preview          & 32B                 & Qwen2.5$^*$        & \na & \na \\ \bottomrule
\end{tabularx}
\caption{Summary of open-weight baseline models. \textit{CP} stands for Continue Pretrain. \textit{SFT} stands for Supervised Fine-Tuning. \textit{GRPO} refers to a variant of PPO, which replaces the value network with the group average \cite{shao2024deepseekmath}. PoT \cite{chenprogram} and TIR \cite{goutora} stand for Program-of-Thought and Tool-Integrated Reasoning, respectively. \textit{PRIME} stands for using ORM as PRM by DPO-like rewards \cite{cui2024process}. \textit{SLM} stands for Selective Language Modeling \cite{lin2024rho}. $^*$ only stands for the same model architecture.}
\label{tab:baseline-summary}
\end{table}
\paragraph{Summary for Open-weight Baselines} The summary of the open-weight baseline models is shown in the following Table~\ref{tab:baseline-summary}. From Table~\ref{tab:baseline-summary}, it is evident that current math-focused LLMs are built on a variety of base models, with \textbf{Mistral} being the most commonly used, followed by \textbf{Llama2}, \textbf{Qwen2-Math}, and \textbf{Deepseek-Math}. Notably, most models utilize data generation and augmentation strategies centered around \textbf{MATH} and \textbf{GSM8K}, which has accelerated the saturation of these benchmarks and highlighted the limitations of current ``mathematical reasoning'' capabilities in LLMs. 

Furthermore, most academia-developed models are often constrained to supervised fine-tuning (SFT) only due to limited computation resources. Recent efforts, such as \textbf{Eurus-2-PRIME} and \textbf{Rho-Math}, have begun to explore advanced pre-training and post-training techniques, which have been (implicitly) validated by companies as effective in enhancing mathematical reasoning. However, the opacity surrounding engineering details in technical reports \cite{openai2023gpt4, shao2024deepseekmath, yang2024qwen2} hinders progress toward genuine mathematical reasoning in LLMs, which is critical for helping math researchers with true mathematical research.

\textbf{Consequently, our proposed conceptual mathematical benchmark, \dataname{}, is a timely and significant contribution to fostering advancements in genuine mathematical reasoning of LLMs.}

\paragraph{Details of Prompts} The used prompts are summarized as follows. We follow the corresponding tokenizers for \textit{chat-template-based prompts} with the instruction as system prompt and statement as user input.

\begin{tcolorbox}[title=\textit{Completion Prompt},top=1mm,bottom=1mm]
\footnotesize
\{statement\}\\
Please reason step by step about whether the above statement is True or False, and put your final answer within $\backslash\backslash$boxed\{\}.
\end{tcolorbox}

\begin{tcolorbox}[title=\textit{Alpaca Prompt},top=1mm,bottom=1mm]
\footnotesize
Below is an instruction that describes a task. Write a response that appropriately completes the request.\\\\
\#\#\# Instruction:\\
\{statement\}\\
Please reason step by step about whether the above statement is True or False, and put your final answer within $\backslash\backslash$boxed\{\}.\\\\
\#\#\# Response:
\end{tcolorbox}

\begin{tcolorbox}[title=\textit{Chat-Template-Based Prompt},top=1mm,bottom=1mm]
\footnotesize
\textless BOS \textgreater system\\
Please reason step by step about whether the statement is True or False, and put your final answer within $\backslash\backslash$boxed\{\}.\textless EOS\textgreater \\
\textless BOS \textgreater user\\
\{statement\}\textless EOS\textgreater 
\end{tcolorbox}

\begin{tcolorbox}[title=\textit{Prompt with Hint},top=1mm,bottom=1mm]
\footnotesize
\{statement\}\\
Please reason by giving examples about whether the above statement is True or False, and put your final answer within $\backslash\backslash$boxed{}.
\end{tcolorbox}

\paragraph{Details of Evaluation Prompt}
The evaluation prompts used are as follows:

\begin{tcolorbox}[title=\textit{Evaluation Prompt},top=1mm,bottom=1mm]
\scriptsize

\# Role \\
You are a professional evaluator responsible for assessing the correctness of students' use of proof by contradiction to solve mathematical proof problems.\\

\#\# Objective \\
- Extract all examples in the reference solution.\\
- Extract all examples presented in the student's proof.\\
- Compare and determine whether the examples provided by the student's proof are semantically equivalent to those in the reference solution.\\

\#\# Workflow \\
1. Read and understand the given mathematical statement and its truth value.\\
2. Extract all examples from the reference solution.\\
3. Extract all examples from the student's proof. If the student does not provide any examples during the proof process, please output ``None" in the \#\#\# $\langle$ Examples in Student's Proof $\rangle$.\\
4. Compare each example provided by the student's proof with the examples in the reference solution, determining whether they convey the same meaning. If they are semantically equivalent, output {\text{CONSISTENCY}}; otherwise, output {\text{UNCONSISTENCY}}.\\

\#\# Constraints \\
- Focus only on counterexamples used within the proof by contradiction, ignoring other proof methods.\\
- Ensure the accuracy of the comparison to avoid misinterpreting the student's intended meaning.\\

\#\# Output Format \\
When a student provides examples in a proof, for instance by stating ``Let's consider an example to illustrate that", what follows may lead to the presentation of an example:
\begin{verbatim}
### <Examples in Reference>
Example 1: ......
Example 2: ......
Example 3: ......

### <Examples in Student's Proof>
Example 1: ......
Example 2: ......

### <Each Result in Examples in Student's Proof>
Example 1: CONSISTENCY
Example 2: UNCONSISTENCY
\end{verbatim}

When the student does not provide any examples:
\begin{verbatim}
### <Examples in Reference>
Example 1: ......
Example 2: ......
Example 3: ......

### <Examples in Student's Proof>
None

### <Each Result in Examples in Student's Proof>
None
\end{verbatim}

\#\# Input
\begin{verbatim}
### Question:
Please judge whether the following statement is True or False:
{statement}

### Reference Answer:
{answer}

### Reference Proof:
{rationale}

### Student Proof:
{model output}
\end{verbatim}
\end{tcolorbox}

\clearpage
\section{Details for Training Data Engineering Framework}
\label{appendix:training_data}
In constructing our training dataset, we utilized GPT to filter and refine the data. Below, we outline the specific prompt designs used in this process:

\paragraph{Data Filtering}
We developed multiple criteria for the LLM to evaluate whether the original data qualified as the counterexample data required for our task. These criteria ensured that the selected data met the conceptual and structural requirements of counterexample reasoning. The specific criteria are as follows:

\begin{tcolorbox}[title=\textit{Data Filtering Prompt},top=1mm,bottom=1mm]
\footnotesize

\# Instruction \\
Please assess the following mathematical proof based on these criteria to determine if it employs counterexamples. \\

\# Criterion \\

\#\# Criterion 1: \\
Does the proof involve assuming the negation of the conclusion and deriving a contradiction from it? This indicates the use of proof by contradiction.\\

\#\# Criterion 2: \\
Is there a specific instance provided that directly refutes a general assertion? This suggests the use of a counterexample. \\

\#\# Criterion 3: \\
Does the explanation include specially chosen cases whose characteristics are essential for illustrating concepts or validating arguments? This points towards the use of special cases. \\

If at least one criterion is satisfied, consider it as meeting the condition and mark it as True. Remember, even if only one criterion applies, please retain this data. \\

\# Input \\
The rationale you need to judge: \\

\{statement\} \\

\{rationale\} \\

\end{tcolorbox}

\paragraph{Data Refinement}

We design prompts for the LLM to rewrite the data, making the training dataset more closely aligned with the benchmark distribution. This refinement process also highlights examples more explicitly, enabling the model to better learn counterexample-based conceptual reasoning. The details of our prompt design are provided below:

\begin{tcolorbox}[title=\textit{Data Refinement Prompt},top=1mm,bottom=1mm]
\footnotesize
\# Task:\\
You are tasked with modifying the proof processes in given rationales. \\

\# Instructions:\\
\#\# Annotate Examples: \\
Identify and annotate any specific examples used within the proof process. \\

\#\# Follow Reference Examples: \\
Use the reference examples and modified before-and-after examples provided below as a guideline for how to modify the text.\\

\#\# Preserve Proof Content: \\
Focus on reducing the length of the proof without altering its core content or logical flow.\\

\# Example References\\
- Field algebra: [Example A]\\

- Field real analysis: [Example B]\\

- Field functional analysis: [Example C]\\

- Field topology: [Example D]\\

\# Modified Before-and-After Examples:\\

\#\# Example 1:\\
\textit{Before}:\\
\textit{After}:\\

\#\# Example 2:
\textit{Before}:\\
\textit{After}:\\

\#\# Example 3:\\
\textit{Before}:\\
\textit{After}:\\

\# Input: \\
\{rationale\}

\end{tcolorbox}

\end{document}